# Recognition of Facial Expressions Based on Salient Geometric Features and Support Vector Machines

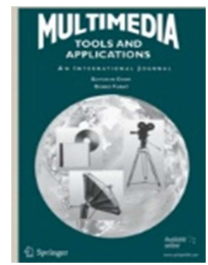

**Deepak Ghimire[1], Joonwhoan Lee[2,*], Ze-Nian Li[3], Sunghwan Jeong[1]**

[1] Korea Electronics Technology Institute, Jeonju-si, Jeollabuk-do 561-844, Rep. of Korea; E-Mails: (deepak, shjeong)@keti.re.kr

[2] Division of Computer Engineering, Chonbuk National University, Jeonju-si, Jeollabuk-do 561-756, Rep. of Korea; E-Mail: chlee@jbnu.ac.kr

[3] School of Computing Science, Simon Fraser University, Burnaby, B.C., Canada; E-Mail: li@cs.sfu.ca

**\*** Author to whom correspondence should be addressed; E-Mail: chlee@jbnu.ac.kr; Tel.: +82-63-270-2406; Fax: +82-63-270-2394.

**Abstract:** Facial expressions convey nonverbal cues which play an important role in interpersonal relations, and are widely used in behavior interpretation of emotions, cognitive science, and social interactions. In this paper we analyze different ways of representing geometric feature and present a fully automatic facial expression recognition (FER) system using salient geometric features. In geometric feature-based FER approach, the first important step is to initialize and track dense set of facial points as the expression evolves over time in consecutive frames. In the proposed system, facial points are initialized using elastic bunch graph matching (EBGM) algorithm and tracking is performed using Kanade-Lucas-Tomaci (KLT) tracker. We extract geometric features from point, line and triangle composed of tracking results of facial points. The most discriminative line and triangle features are extracted using feature selective multi-class AdaBoost with the help of extreme learning machine (ELM) classification. Finally the geometric features for FER are extracted from the boosted line, and triangles composed of facial points. The recognition accuracy using features from point, line and triangle are analyzed independently. The performance of the proposed FER system is evaluated on three different data sets: namely CK+, MMI and MUG facial expression data sets.

## 1. Introduction

The tracking and recognition of facial activities from still images or video sequences has attracted great attention in computer vision field. Among them recognition of facial expression has been an active research topic since last decade. Facial expressions are among the most universal forms of body language. A facial expression is one or more motions, or positions of the muscles beneath the skin of the face. These movements convey the emotional state of an individual to observers. Psychological research has shown that facial expressions are the most expressive way in which human display emotions [1]. In general, researchers divided facial expressions into six basic categories: anger, disgust, fear, happiness, sadness, and surprise; which are also called primary emotions [2]. Advanced emotions include frustration and confusion, sentiments including positive, negative, and neutral, composite of two or more emotions, facial action units etc.

In the digital age, it is no secret that social relationships are changing. With all advantages that our digital devices have brought us, they are also affecting our ability to empathize with others. Recent research shows that due to excessive use of digital devices young people are losing their ability to read other people emotions or feeling [3]. Therefore it is important to recognize emotions via facial expressions accurately and in real time. On the other hand, facial emotion recognition has applications in human-computer interaction, clinical studies, advertising, action recognition for computer games, etc.

An automatic FER system generally consists of three steps [4]: (a) accurate localization of face in an image or video, (b) facial feature extraction and representation, and finally (c) recognition of facial expression using feature classification. In this paper we focus on the study of salient geometric feature extraction for recognizing the six basic prototypical facial expressions. Fig.1 shows the overall block diagram of the proposed FER system. As shown in fig. 1, at first, face detection, feature point initialization, and tracking is performed. Viola and Jones Haar like feature based AdaBoost scheme [5] is used for face and eye detection, whereas EBGM [6] and KLT tracker [7] is used for feature point initialization and tracking in consecutive video frames, respectively. Face graph normalization scheme is proposed to bring all face graphs in standard shape before feature selection and extraction. Three different geometric features are extracted. 1) Single facial points coordinate displacements feature, 2) two points are considered at once to form line features, and 3) three facial points are considered at once to form triangle type features. Prominent line and triangle are selected using multi-class AdaBoost before feature extraction. Detail of this procedure will be discussed in section 3. Finally facial expressions are recognized using SVMs learned on points, lines, and triangles based geometric features, independently. We analyze different types of geometric feature extraction and present the recognition results in different data sets.

The main contributions of this paper are summarized as follows:

1. We propose a fully automatic sequence based FER system using salient geometric feature representations.
2. We study facial geometric feature in three different forms (point, line and triangle); their representation power for discriminating basic FER are compared, and validated using publicly available three different FER data sets.

3. We show that the triangle based representation outperforms both line and point based representation, whereas line based representation outperforms point based feature representation. Therefore our study proves that, not only the facial feature movement over time but also the inter-relation between facial features movements within a face is important in discriminating facial expressions.
4. We conduct extensive FER experiments on three widely used facial expression data sets to demonstrate the efficiency of our proposed method. Experimental results show that our method is superior to most state-of-the-art FER systems.

Rest of this paper is organized as follows. A brief review of the work in the field of FER is given in section 2. Face detection, facial point initialization, tracking, and normalization of face graph as well as different types of geometric feature extraction is described in section 3. Section 4 describes the analysis and selection of different geometric feature from the tracking result of dense set of facial points. The experimental setup and dataset description is given in section 5. Experimental results on different publicly available benchmark facial expression data sets are presented in section 6. Finally, conclusion of the proposed FER system is given in section 7.

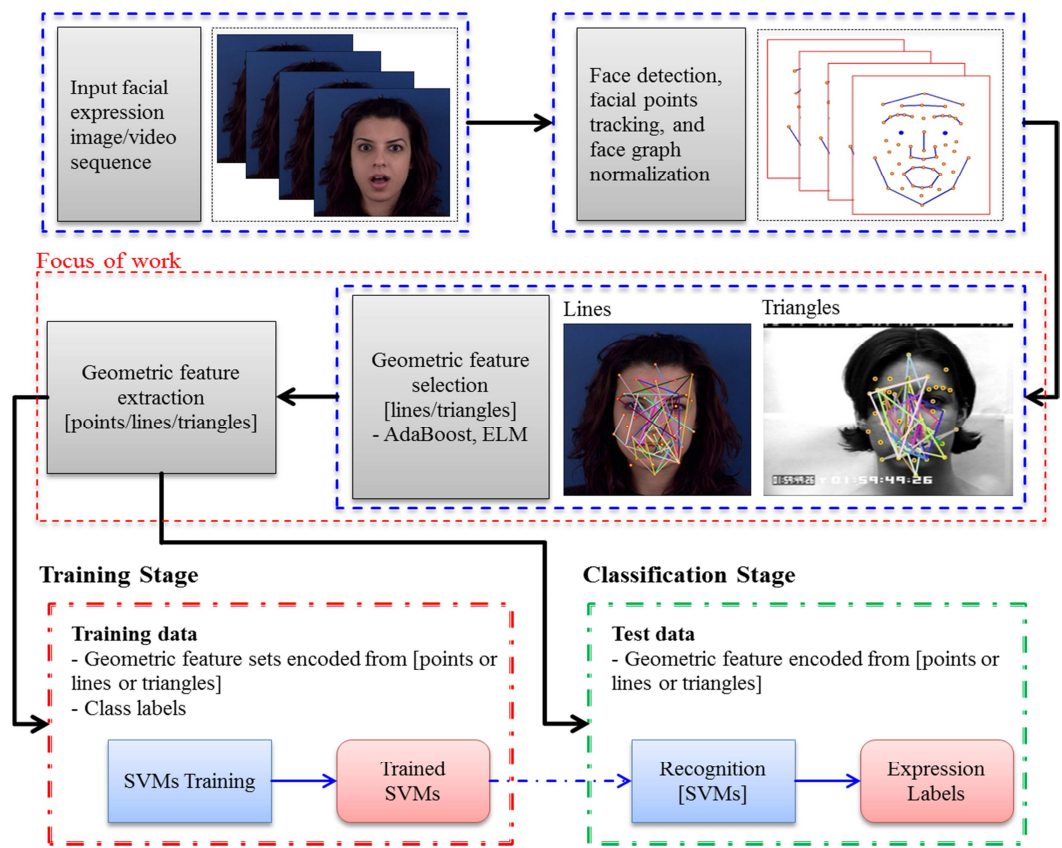

**Fig. 1.** Architecture of the proposed facial expression recognition system

## 2. Related Work

Several researchers have presented review on FER; among them early reviews can be found in [8, 9, 10], whereas recent reviews can be found in [11, 12]. First review was made in 1994 by Samal and Iyenger [8], followed by [9], [10] in 2000 and 2003, respectively. In [11], survey of affect recognition methods including audio, visual and spontaneous expressions is made, which was published in 2009. It covers discussion of emotion perception from psychological perspective, examination of available approaches for solving problem of machine understanding of human affective behavior, discussion on collection and availability of emotion training dataset, and also outlines the scientific and engineering challenges to advancing human affect sensing technology. Recently, in 2012, meta-review of FER and analysis challenge is presented by Valster et al. [12], in which the focus is on clarifying how far the field has come, identifying new goals, and providing baseline results regarding facial emotion recognition and analysis.

The approaches reported for FER can be classified into two main categories, a) template-based and b) feature-based. The template-based methods use 2-D or 3-D facial models as templates for expression information extraction. The feature-based methods use appearance-based features or geometry-based features for expression information extraction. Geometry-based features describe the shape of the face and its components such as the mouth or the eyebrow, whereas appearance-based features describe the texture of the face caused by expression.

Among the appearance-based features, local binary pattern (LBP) is widely used recognizing facial expressions [13-17]. Similarly, local Gabor binary pattern [16], histogram of orientation gradient [18], Gabor wavelets representation [17], scale invariant feature transform (SIFT) [19], non-negative matrix factorization (NMF) based texture features [20, 21], linear discriminant analysis (LDA) [22], independent component analysis (ICA) [22] etc., are also widely used appearance-based feature for the recognition of facial expressions.

Most geometric feature-based approaches use the active appearance model (AAM) or its variation, to track a dense set of facial points [23, 24]. Alternatively, EBGM algorithm, KLT tracker etc., are also used for facial key point detection or tracking [25, 26]. The locations of these facial landmarks are then used in different ways to extract facial features regarding shape of the face, or movement of facial key points as the expression evolves. Kotisa *et al*. [26] used geometric displacement of certain selected candid nodes, defined as the differences of the node coordinates between the first and the greatest facial expression intensity frames, as geometric features for recognition of six basic facial expressions. Sung and Kim [27] introduced Stereo AAM, which improves the fitting and tracking of standard AAMs using multiple cameras to model the 3-D shape and rigid motion parameters. Active shape model (ASM) is used in [28] for modeling and tracking facial key points and the facial expressions are recognized on a low-dimensional expression manifold. Pose invariant FER based on a set of characteristic facial points extracted using AAMs is presented by Rudovic and Pantic [29]. A coupled scale Gaussian process regression (CSGPR) model is used for head-pose normalization. Ghimire and Lee [25] used tracking result of 52 facial key points modeled in the form of point and line features for the recognition of facial expressions. The key geometric features are selected based on AdaBoost and dynamic time warping (DTW) algorithm. Recently, in [30] and [31], authors also utilized geometric features for the recognition of facial expressions. In [30], facial activities are characterized by three levels. First, in the bottom level, facial feature point are tracked using ASM, in the middle level, facial action units are defined, and finally facial expressions are represented based on detected action units. Saeed *et al.* [31] use only eight facial key points in order to model geometric structure of face from single face image for the recognition of facial expressions.

Several classifiers have also been investigated to build recognition module for facial expressions. Therefore, FER techniques can also be categorized according to recognition modules. The most common recognition modules are support vector machines (SVMs), hidden Markov models (HMM), Gaussian mixture models (GMM), dynamic Baysian networks (DBN) etc. Among them [13, 16, 17, 24, 25, 26, 29, 31] use SVM, HMM is used in [32, 37, 38], GMM is utilized by [35, 36], whereas [30, 33] uses DBN. Recently, sparse representation classification (SRC), which is a very successful face recognition technique [34], is also used for FER [17]. In SVM, the probability is calculated using N-fold cross validation technique, in other words, there is no direct probability estimation in SVM. Therefore in order to recognize facial expressions from video, the temporal information should be embedded in feature extraction process. GMM is sensitive to noise and cannot model fast variation in the consecutive frames. HMM are mostly used to handle the sequential data when frame level features are used which has the advantage over SVM, GMM like classifiers.

## 3. Geometry-based Facial Feature Extraction

Facial feature extraction attempts to find the most appropriate representation of the face image for recognition. In geometric feature-based system, the facial points in a single image or in image sequences are used in different ways to form feature vector for recognition of facial expressions. For example, the distance between feature points and the relative sizes of the major face components are computed to form a feature vector. The feature points can also form a geometric graph representation of the face. Using geometric features have their own advantage and disadvantages. The difficulty in geometric feature based approach is to initialize and track facial feature points accurately and in real time. If there is error in feature point initialization and tracking, the error gets accumulated in the geometric feature extraction process. Image resolution, head pose, eyeglass, presence of beard etc. could also affect the feature point initialization and tracking process. But once the feature points are initialized and tracked accurately, the geometric features extracted from the tracking result are robust to variation in scale, size, head orientation, texture of the face due to age variation etc.

In this section we will present the method for facial feature point initialization and tracking. Different type of geometric feature extraction technique, as well as feature selection technique to find the most discriminant geometric features for the recognition of facial expressions will be studied. The geometric features are extracted based on point, line and triangle composed of facial key points in the video sequence.

### 3.1. Facial Feature Point Tracking and Graph Normalization

In the proposed method, the facial points are initialized and tracked automatically. The feature point initialization is performed using EBGM algorithm. The tracking in consecutive frames is performed using KLT tracker. Finally, the face graph is normalized in such a way that for each facial expression sequence, the vertex of the initial face graph starts from same position and evolves according to the movements of facial feature points as some particular facial emotion evolves over time.

We use EBGM algorithm, which was implementation by Colorado State University (CSU) as a baseline algorithm for the comparison of face recognition algorithms [39], for facial feature point initialization. Facial feature point localization in a novel imagery has two steps. First, the locations of the new feature point is estimated based on the known locations of other feature points in the image; second, the estimate is refined by comparing the Gabor jet extracted from that image in the approximate locations and the jets extracted from the same positions in the model images. In order to start the feature point localization process, the approximate locations of two eyes are detected using Haar-like feature based object detection algorithm [5].

Once the facial feature points are automatically initialized using EBGM algorithm we use pyramidal variant of well-known KLT tracker for tracking the 52 facial feature points in consecutive frames. The KLT algorithm tracks a set of feature points across the video frames. The algorithm tracks the facial feature points in the image sequence containing the formation of a dynamic human facial expression from the neutral state to the fully expressive one. KLT tracking is faster as compared to the EBG using Gabor filter based tracking algorithm used in [25]. Fig. 2 shows the result of facial feature point tracking using KLT tracker.

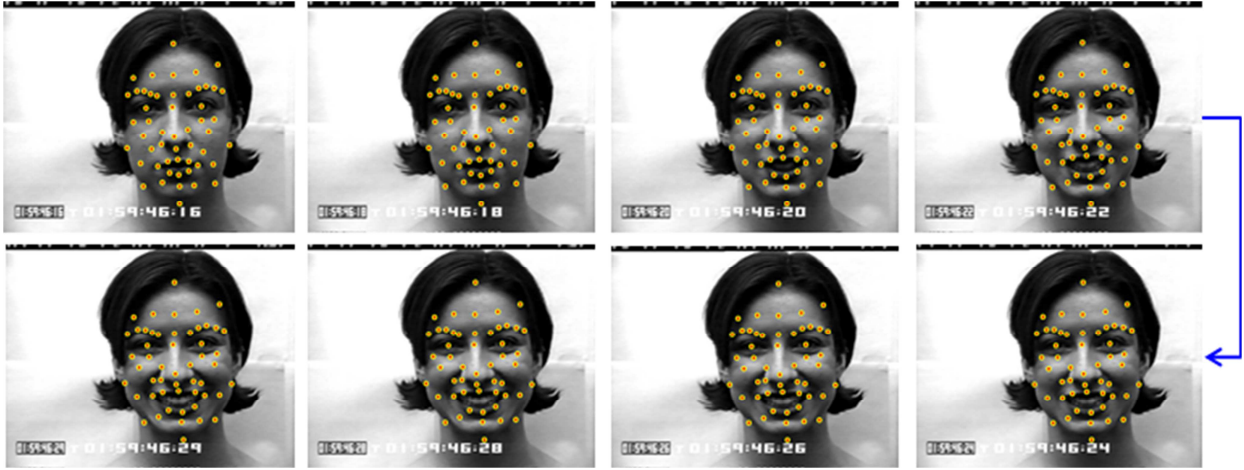

**Fig. 2.** An example of facial feature point tracking in happy facial expression sequence using KLT tracker.

Face graph normalization brings each face points to the uniform coordinate position in the first frame of the video shot, and as the expression evolves, the landmarks are displaced accordingly. Let us suppose $(x_l, y_l)_i^k$ denotes the $i^{th}$ feature point position in the $l^{th}$ frame of the $k^{th}$ facial expression sequence in the database. Tracking result of a single landmark is denoted by $S_i^k$, and is defined by Eq. (1).

$$S_i^k = \left[ (x_0, y_0)_i^k, (x_1, y_1)_i^k, ...., (x_N, y_N)_i^k \right] \quad (1)$$

where, $N$ is the number of frames in an expression sequence.

An average feature point position corresponding to each feature point is computed by averaging feature points in neutral face images, i.e., first frame in the video shot. Suppose $(\mu_{x0}, \mu_{y0})_i$ denotes the average key point position of the $i^{th}$ key point in the first frame of the expression sequence. Suppose $(\delta_{x0}, \delta_{y0})_i^k$ denotes the displacement of the $i^{th}$ key point in the first frame of the $k^{th}$ expression sequence, with respect to the average key point position:

$$(\delta_{x0}, \delta_{y0})_i^k = (\mu_{x0} - x_0, \mu_{y0} - y_0)_i^k \qquad (2)$$

Now the key point displacement described by Eq. 2 is added to the key point positions in every frame of the expression sequence. The transformed result of key point tracking is now denoted by $S'^k_i$ and is defined as:

$$S'^k_i = \left[ (x_0 + \delta_{x0}, y_0 + \delta_{y0})_i^k, (x_1 + \delta_{x0}, y_1 + \delta_{y0})_i^k, ..., (x_N + \delta_{x0}, y_N + \delta_{y0})_i^k \right] \qquad (3)$$

Fig. 3 shows the result of the facial feature tracking and corresponding result after graph normalization. Note that graph is also scaled in order to make uniform size. Also, note that, the lines connecting two feature point are used just to make face like appearance.

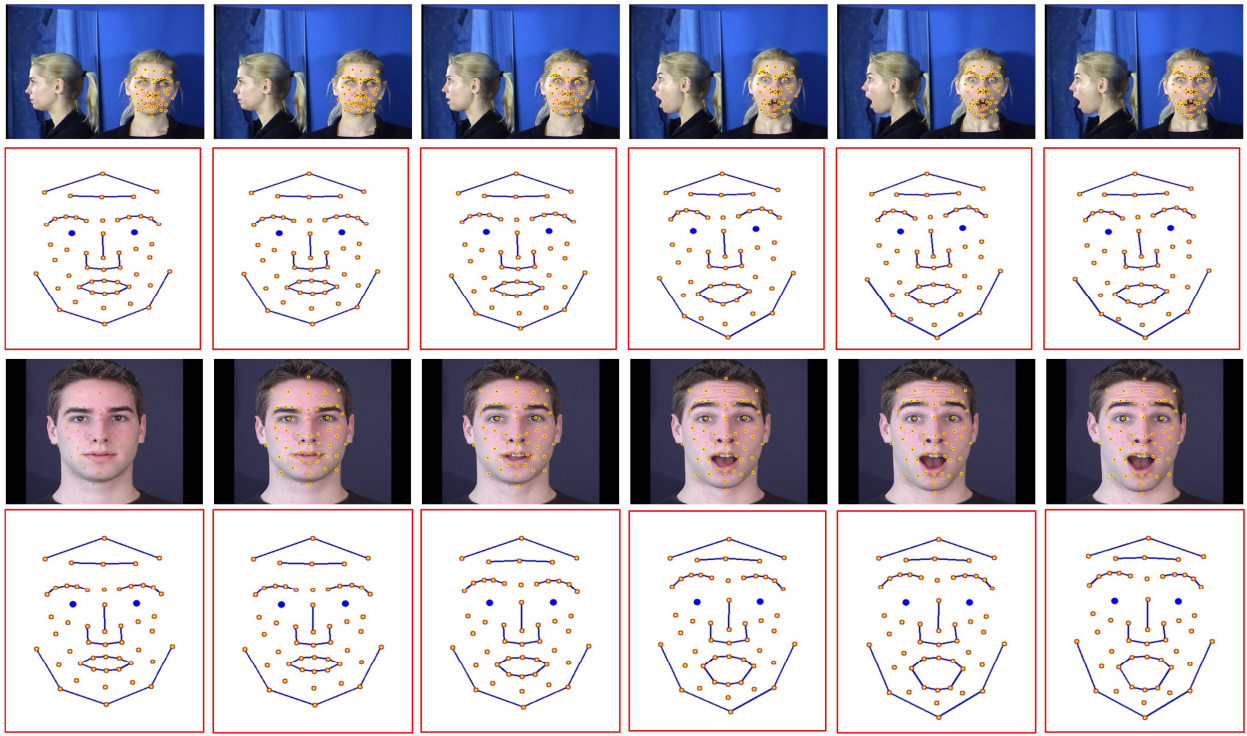

**Fig. 3.** Example of facial feature point tracking and corresponding result after normalization for two surprise facial expression sequences from MMI database.

*3.2. Point Based Geometric Features*

Suppose $(x', y')$ is the normalized key point coordinate position in the face graph, and let us rewrite Eq. 3 in the following form:

$$S'^k_i = \left[ (x'_0, y'_0)_i^k, (x'_1, y'_1)_i^k, ..., (x'_N, y'_N)_i^k \right] \qquad (4)$$

The number of frames in different video shots of facial expression can be different. To make feature extraction and feature selection process easy, the feature point tracking result is resized to fixed length using linear interpolation. In our experiment the sequence is resampled into $N = 10$ frames.

The feature point displacement in each frame with respect to the first frame is calculated. Suppose $(\Delta x'_l, \Delta y'_l)^k_i$ denotes the difference between the $i^{th}$ landmark in the $l^{th}$ frame, and $i^{th}$ landmark in the first frame of the $k^{th}$ facial expression sequence in the database.

$$(\Delta x'_l, \Delta y'_l)^k_i = (x'_l - x'_0, y'_l - y'_0)^k_i \qquad (5)$$

Eq. (6) defines all the displacements of the $i^{th}$ feature point in the $k^{th}$ sequence.

$$\Delta S'^k_i = \left[ (\Delta x'_1, \Delta y'_1)^k_i, (\Delta x'_2, \Delta y'_2)^k_i, \ldots, (\Delta x'_N, \Delta y'_N)^k_i \right] \qquad (6)$$

### 3.3. Line Based Geometric Features

The geometric feature extracted in the form of Eq. (6) considers only the tracking result of individual feature points. The movements of key points as the particular facial expression evolves are not independent, i.e., there is definite relationship between the movements of facial key points. In order to capture this information in the feature, pair of feature points is considered at a time, and then features are extracted as components of line. The Euclidian distance and the base angle connecting pair of facial key points within a frame are calculated as a line based geometric features.

Suppose $(d_l, \theta_l)^k_{i,j}$ denotes the Euclidian distance and angle between $i^{th}$ and $j^{th}$ pair of key points in the $l^{th}$ frame of the $k^{th}$ facial expression sequence.

$$d^k_{l,(i,j)} = \sqrt{(x^k_{l,i} - x^k_{l,j})^2 + (y^k_{l,i} - y^k_{l,j})^2} \qquad (7)$$

$$\theta^k_{l,(i,j)} = \arctan\left( \frac{y^k_{l,i} - y^k_{l,j}}{x^k_{l,i} - x^k_{l,j}} \right) \qquad (8)$$

Let us denote the calculated sequence of distances and angle by $L^k_{i,j}$, and defined by Eq. (9):

$$L^k_{i,j} = \left[ (d_0, \theta_0)^k_{i,j}, (d_1, \theta_1)^k_{i,j}, \ldots, (d_N, \theta_N)^k_{i,j} \right] \qquad (9)$$

Now, the obtained distance and angle between the pair of landmarks are subtracted from the corresponding distance and angle in the first frame of the video shot. Suppose $(\Delta d_l, \Delta \theta_l)^k_{i,j}$ denotes the change in distance and angle between $i^{th}$ and $j^{th}$ pair of key points in the $l^{th}$ frame, with respect to the first frame of the $k^{th}$ video shot, which is defined as:

$$(\Delta d_l, \Delta \theta_l)^k_{i,j} = (d_l - d_0, \theta_l - \theta_0)^k_{i,j} \qquad (10)$$

Finally, the line based geometric feature extracted from the image sequence is defined as follows:

$$\Delta L^k_{i,j} = \left[ (\Delta d_1, \Delta \theta_1)^k_{i,j}, (\Delta d_2, \Delta \theta_2)^k_{i,j}, \ldots, (\Delta d_N, \Delta \theta_N)^k_{i,j} \right] \qquad (11)$$

### 3.4. Triangle Based Geometric Features

Here, three facial landmarks are considered at a time and features are extracted in the form of components of triangle. The information regarding movement of facial key points and relationship between them when some facial expression evolves over time can be captured well by considering three facial key points at a time as compared to two facial key points. Triangle components in the $l^{th}$ frame are subtracted with the triangle components in the first frame of the video sequence as shown in fig. 4.

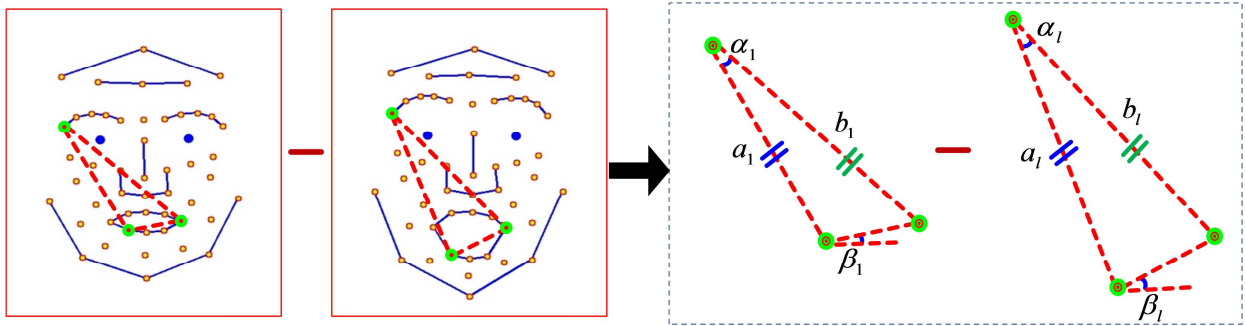

**Fig. 4.** Difference in components of two triangle used as features. Vertex of each triangle corresponds to the facial key point in the two frames of the video sequence.

Suppose $(a_l, b_l, \alpha_l, \beta_l)^m_{i,j,k}$ denotes the two side lengths, an included angle, and the base angle of the triangle composed of $i^{th}$, $j^{th}$ and $k^{th}$ key points of the face graph in the $l^{th}$ frame of the $m^{th}$ facial expression sequence.

Now let us denote the calculated sequence of triangle components as shown in fig. 4 by $T^m_{i,j,k}$, and defined as in Eq. (12).

$$T^m_{i,j,k} = \left[(a_0, b_0, \alpha_0, \beta_0)^m_{i,j,k}, (a_1, b_1, \alpha_1, \beta_1)^m_{i,j,k}, ...,(a_N, b_N, \alpha_N, \beta_N)^m_{i,j,k}\right] \quad (12)$$

The obtained components of triangle in the sequence are now subtracted from the corresponding components of triangle in the first frame of the video sequence. Suppose $(\Delta a_l, \Delta b_l, \Delta \alpha_l, \Delta \beta_l)^m_{i,j,k}$ denotes the difference between components of triangle in the $l^{th}$ frame of the video shot and the components of the corresponding triangle in the first frame of the video sequence, which is defined as:

$$(\Delta a_l, \Delta b_l, \Delta \alpha_l, \Delta \beta_l)^m_{i,j,k} = (a_l - a_0, b_l - b_0, \alpha_l - \alpha_0, \beta_l - \beta_0)^m_{i,j,k} \quad (13)$$

Finally, triangle based geometric feature extracted from the image sequence are defined as follows:

$$\Delta T^m_{i,j,k} = \left[(\Delta a_1, \Delta b_1, \Delta \alpha_1, \Delta \beta_1)^m_{i,j,k}, (\Delta a_2, \Delta b_2, \Delta \alpha_2, \Delta \beta_2)^m_{i,j,k},...,(\Delta a_N, \Delta b_N, \Delta \alpha_N, \Delta \beta_N)^m_{i,j,k}\right] \quad (14)$$

Suppose there are $N$ frames in the sequence, then the feature vector is composed of $(N-1) \times 4$ components, i.e., if $N = 11$, feature dimension for a sequence extracted from the single triangle will be $(11-1) \times 4 = 40$.

## 4. Features Selection using Multi-class AdaBoost with ELM

The geometric features are extracted in the form of components of lines and triangle. In total there are 52 facial key points. According to the combination principle with 52 facial points $_{52}C_2 = 52!/(2!(52-2)!) = 1326$ and $_{52}C_3 = 52!/(3!(52-3)!) = 22100$ unique lines and triangles are possible. If we use all of them in order to extract features for classification, the dimension of the feature will be large. In feature extraction process each point, line, and triangle in the face graph are represented by Eq. (6), Eq. (11), and Eq. (14), respectively. Let us call these equations as feature vector, because in this paper feature selection means selection of lines or triangles whose components represents the discriminative feature for the recognition of facial expressions. Among large number of feature vectors only the small subset will provide discriminative information for recognition of facial expressions. Our goal is to find subset of lines and triangles using some feature selection scheme. Here we use feature selective AdaBoost algorithm in combination with ELM.

### 4.1. Extreme Learning Machine

Gradient based learning algorithms are very slow and may easily converge to local minima. They also require many iterative learning steps in order to obtain better learning performance. ELM, a fast learning algorithm for single-layer feed-forward neural networks (SLFNs) proposed by Huang et al. [40], solves the gradient-based learning algorithm by analytically calculating the optimal weights of the SLFN. First, the weights between the input layer and hidden layer are randomly selected and then the optimal values for the weights between hidden layer and output layer are determined by calculating the linear matrix equations.

In summary, ELM algorithm can be written as follows:

**Algorithm 1**. Summary of extreme learning machine (ELM) algorithm.

*Given a training set* $\{(x_i, t_i) | x_i \in R^n, t_i \in R^m, i = 1,...,N\}$, *hidden node output function* $g(w,b,x)$, *and number of hidden nodes L,*
  a. *Randomly assign hidden node parameters* $(w_i, b_i)$, $i = 1,...,L$.
  b. *Calculate the hidden layer output matrix H.*
  c. *Calculate the output weights* $\beta$ : $\beta = H^\dagger T$.

*where* $H^\dagger$ *is the Moore-Penrose generalized inverse of hidden layer output matrix H.*

Multi-class AdaBoost algorithm (Algorithm 2) is used to select the salient lines and triangles. ELM is used as a weak classifier in AdaBoost algorithm. ELM itself is not a weak classifier, but in the proposed system, in terms of feature it is treated as a weak classifier, i.e., ELM will be trained using feature extracted from single line or single triangle. The reason behind selecting ELM as a weak classifier is that it is a very fast learning algorithm, and can be trained almost in real time. In the proposed feature selection scheme we have to train 1326 ELMs for line feature selection and 22100 ELMs for triangle feature selection.

### 4.2. Feature Selective Multi-class AdaBoost

The AdaBoost learning algorithm proposed by Freud and Schapire [41], in its original form, is used to boost the classification performance of a simple learning algorithm. In our system, a variant of multi-class AdaBoost proposed by Jhu et al. [42] is used to select the lines or triangles from which features will be extracted for FER.

**Algorithm 2:** Multi-class AdaBoost learning algorithm. *M* hypothesis are constructed, each using a single feature vector. The final hypothesis is a weighted linear combination of *M* hypothesis.

---

1. Initialize the observation weights $w_{1,i} = 1/n, i = 1, 2, ..., n$

2. For $m = 1$ to $M$:

   a. Normalize the weights $w_{m,i} \leftarrow w_{m,i} / \sum_{j=1}^{n} w_{m,j}$

   b. Select the best weak classifier with respect to the weighted error
   $$err^{(m)} = \min_f \sum_{i=1}^{n} w_i . I(c_i \neq T(x_i, f)) / \sum_{i=1}^{n} w_i$$

   c. Define $T^{(m)}(x) = T(x, f_m)$ where $f_m$ is the minimize of $err^{(m)}$

   d. Compute $\alpha^{(m)} = \log \dfrac{1 - err^{(m)}}{err^{(m)}} + \log(K - 1)$.                                (15)

   e. Update the weights: $w_{m,i} \leftarrow w_{m,i} . \exp\left(\alpha^{(m)} . I(c_i \neq T^{(m)}(x_i))\right), i = 1, ..., n$

3. The final strong classifier is: $C(x) = \arg\max_k \sum_{m=1}^{M} \alpha^{(m)} . I(T^{(m)}(x) = k)$

---

Algorithm 2 shows the variant of multi-class AdaBoost learning algorithm proposed in [42], in which they refer their algorithm as SAMME – Stagewise Additive Modeling using a Multi-class Exponential loss function. Weak classifier $T(x, f)$ in our system is trained ELM network using features extracted from single line or triangle. Note that we performed line and triangle selection experiment independently. But the process used for feature selection is same. The multi-class AdaBoost algorithm given in Algorithm 2 is similar to AdaBoost, with the major difference in Eq. (15). Now in order for $\alpha^{(m)}$ to be positive, we only need $(1 - err^{(m)}) > 1/K$, where K is the number of classes, or the accuracy of each weak classifier to be better than random guessing rather than 1/2.

The feature extraction based on the tracking result of individual landmark is simple process. It gives the maximum displacement of feature point in four directions as some expression evolves over time. But feature selection is only applied for line and triangle based feature extraction process. Before creating the feature vector for expression recognition, the feature selection process selects those line and triangle which carry most of information for discriminating six basic facial expressions. Fig. 5 and fig. 6 shows the most discriminative lines and triangle features in three different data sets. Note that line and triangle feature selection is performed independently. Mostly, the selected lines or triangles are composed of landmarks from the eyebrow, mouth and nose area. In most cases, at least one of the vertexes of the triangle is from the eyebrow or mouth or nose region.

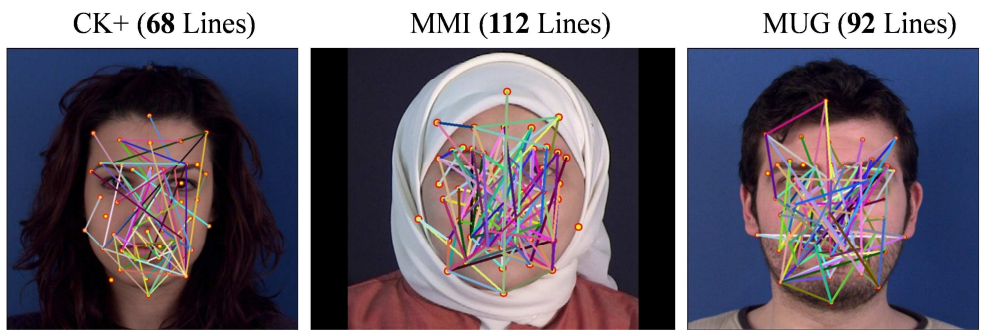

**Fig. 5.** Set of lines selected using multi-class AdaBoost with ELM in three different data sets (left to right: CK+, MMI, and MUG data sets).

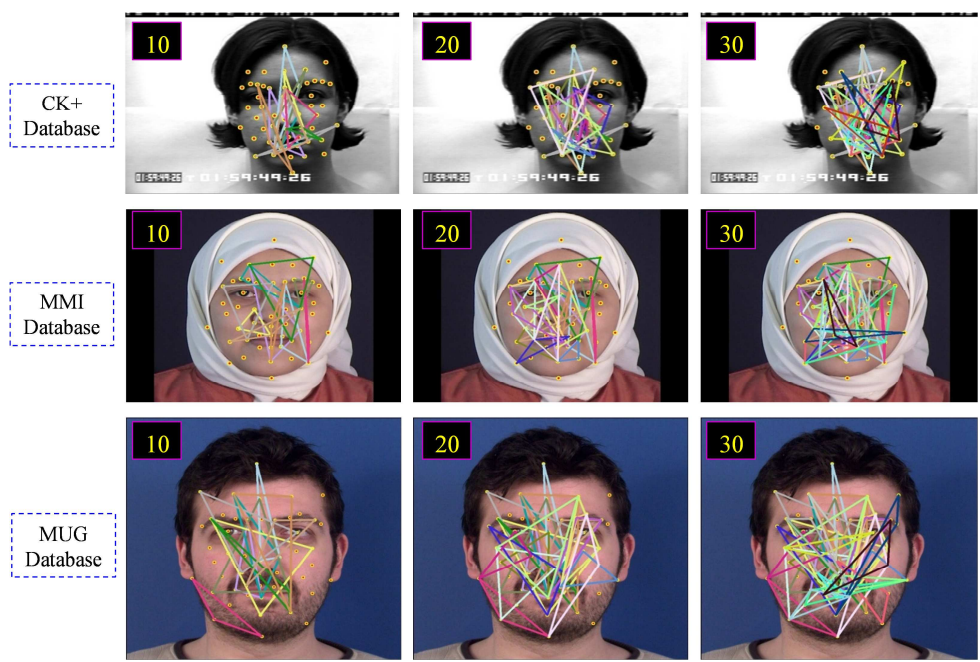

**Fig. 6.** First 10, 20, and 30 triangular features selected using multi-class AdaBoost with ELM. First row: CK+ data set, second row: MMI data set, and third row: MUG data set.

### 5. Experimental Setup and Data sets Description

In order to access the reliability of the proposed FER approach, the performance of the proposed FER system is evaluated on three different databases: extended Cohn-Kanade (CK+) facial expression dataset [43], M&M Initiative (MMI) dataset [44], and Multimedia Understanding Group (MUG) dataset [45]. These dataset consists of facial expression image sequence or videos which starts from natural frame and evolves to peak facial expression intensity.

The most common approach for testing the generalization performance of a classifier is the *K*-fold cross validation approach. A ten-fold cross validation was used in order to make maximum use of the available data, and produce averaged classification accuracy results. The classification accuracy is the average accuracy across all ten trials. To get better picture of the recognition accuracy of each expression type, the confusion matrices are given. The diagonal entries of the confusion matrix are the rates of facial expressions that are correctly classified, while the off-diagonal entries correspond to misclassification rates.

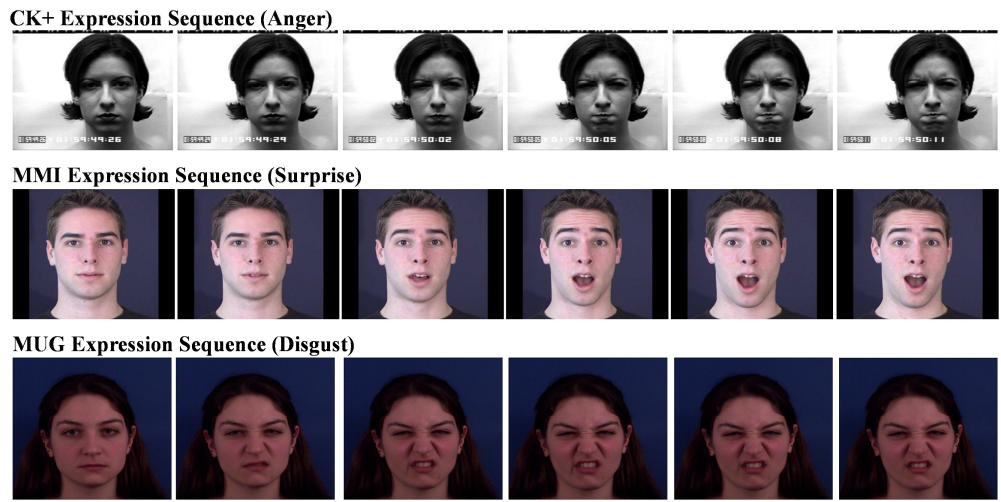

**Fig. 7.** Example of facial expression sequence from three different data sets

SVM is a well-known classifier for its generalization capability. SVM classifiers maximize the hyper plane margin between classes. In our experiment, we use a publicly available implementation of SVM, *libsvm* [46], in which we used radial basic function (RBF) kernel. The optimal parameter selection is performed based on the grid search strategy [47].

A brief introduction of three different data sets used in this paper is given below.

**Extended Cohn-Kanade (CK+) dataset:** The extended Cohn-Kanade (CK+) dataset [43] was used for FER in six basic facial expression classes (anger, disgust, fear, happiness, sadness, and surprise). This database consists of 593 sequences from 123 subjects. The image sequence varies in duration (i.e., 7 to 60 frames), and incorporates the onset (which is also the neutral face) to peak formation of the facial expression. Image sequences from neutral to target display were digitized into 640×480 or 640×490 pixel arrays. Only 327 of the 593 sequences have a given emotional class. This is because these are the only ones that fit the prototypic definition. For the evaluation of proposed FER system, 315 sequences of the dataset are selected from the database. Fig. 7 (first row) shows an example of the facial expression sequence from CK+ dataset.

**M&M Initiative (MMI) dataset:** The MMI face dataset [44] contains more than 1500 samples of both static images and image sequences of faces in frontal and in profile view displaying various facial expressions of emotion, single AU activation, and multiple AU activation. It not only contains posed but also contains spontaneous expressions of facial behavior. There are approximately 30 profile-view and 750 dual-view facial-expression video sequences. All video sequences have been recorded at a rate of 24 frames per second using a standard PAL camera. It includes 19 different face of students and research staff members of both sexes (44% female), ranging in age from 19 to 62, having either a European, Asian, or South American ethnic background. Total of 203 facial expression video sequences are chosen for the evaluation of the proposed FER systems. Fig. 7 (second row) shows an example of the facial expression sequence from MMI database.

**Multimedia Understanding Group (MUG) dataset:** Image sequences in MUG dataset [45] begin and end at neutral state and follow the onset, apex, offset temporal pattern. For each of the six basic expressions a few image sequences of various lengths are recorded. Each image sequence contains 50 to 160 images. Prior to the recordings, a short tutorial about the basic emotions was given to the subjects. The recordings of 77 subjects are available to researchers and the number of the available sequences counts up to 1462. The database includes 86 subjects with Caucasian origin and age between 20 and 35 years. There are 35 female and 51 males with or without beard. The recorded sequence consist of images saved in high quality lossy JPEG format, with a resolution of 896×896 pixels and a size ranging from 240 to 340 KB. Image sequences of 52 subjects and the corresponding annotation are available publically via the internet. In the proposed system 325 sequences are selected for the experiment. Fig. 7 (last row) shows an example of the facial expression sequence from MUG database.

Table 1 shows the number of facial expression images/video sequences for each expression from each dataset used in this paper for the experimentation of the proposed FER system.

**Table 1.** Number of facial expression images/video sequences in three different data sets.

| Dataset/Expression | Anger | Disgust | Fear | Happiness | Sadness | Surprise | Total |
|---|---|---|---|---|---|---|---|
| CK+ | 44 | 62 | 27 | 69 | 32 | 81 | **315** |
| MMI | 32 | 30 | 28 | 42 | 31 | 40 | **203** |
| MUG | 56 | 55 | 51 | 55 | 51 | 56 | **324** |

## 6. Experimental Results and Discussion

*6.1. Facial Expression Recognition using Point based Features*

In this paper basically three different types of facial geometric features are used individually for the recognition of facial expressions. As explained in section 3.2, the point based feature refers to the geometric features which are individual facial feature point displacement in four possible directions. The feature for SVM classification from $i^{th}$ facial key point of the $k^{th}$ facial expression sequence is explained as follows:

$$\Delta x_{i,\max}^k = \max\left(\Delta x_{1,i}^k, \Delta x_{2,i}^k, ... \Delta x_{N,i}^k\right)$$
$$\Delta x_{i,\min}^k = \min\left(\Delta x_{1,i}^k, \Delta x_{2,i}^k, ... \Delta x_{N,i}^k\right)$$
$$\Delta y_{i,\max}^k = \max\left(\Delta y_{1,i}^k, \Delta y_{2,i}^k, ... \Delta y_{N,i}^k\right)$$
$$\Delta y_{i,\min}^k = \min\left(\Delta y_{1,i}^k, \Delta y_{2,i}^k, ... \Delta y_{N,i}^k\right)$$

Total of 52 facial key points are tracked, therefore the dimensionality of point based feature is $52 \times 4 = 208$. The average recognition accuracy using point based feature with ten-fold cross validation is 96.37%, 67.64%, and 91.41% in CK+, MMI, and MUG facial expression data sets, respectively. Table 2, 3 and 4 show the corresponding confusion matrices labeled with point based feature representation along with line and triangle based features.

**Table 2.** Confusion matrix for FER in percentages using SVM classifier with different types of geometric features representation in CK+ dataset.

| % | Feature Representation | Anger | Disgust | Fear | Happiness | Sadness | Surprise |
|---|---|---|---|---|---|---|---|
| **Anger** | Point | **97.50** | 0 | 0 | 0 | 2.5 | 0 |
| | Line | **97.50** | 0 | 0 | 0 | 2.5 | 0 |
| | Triangle | **97.50** | 0 | 0 | 0 | 2.5 | 0 |
| **Disgust** | Point | 3.33 | **96.67** | 0 | 0 | 0 | 0 |
| | Line | 3.33 | **96.67** | 0 | 0 | 0 | 0 |
| | Triangle | 3.33 | **96.67** | 0 | 0 | 0 | 0 |
| **Fear** | Point | 0 | 0 | 92 | 4 | 0 | 4 |
| | Line | 0 | 0 | 92 | 4 | 0 | 4 |
| | Triangle | 0 | 0 | **96** | 0 | 0 | 4 |
| **Happiness** | Point | 0 | 0 | 0 | **100** | 0 | 0 |
| | Line | 0 | 0 | 0 | **100** | 0 | 0 |
| | Triangle | 0 | 0 | 0 | **100** | 0 | 0 |
| **Sadness** | Point | 6.67 | 0 | 0 | 0 | 93.33 | 0 |
| | Line | 3.33 | 3.33 | 0 | 0 | 93.33 | 0 |
| | Triangle | 3.33 | 0 | 0 | 0 | **96.67** | 0 |
| **Surprise** | Point | 0 | 0 | 1.25 | 0 | 0 | 98.75 |
| | Line | 0 | 0 | 0 | 0 | 0 | **100** |
| | Triangle | 0 | 0 | 0 | 0 | 0 | **100** |

**Table 3.** Confusion matrix for FER in percentages using SVM classifier with different types of geometric features representation in MMI dataset.

| % | Feature Representation | Anger | Disgust | Fear | Happiness | Sadness | Surprise |
|---|---|---|---|---|---|---|---|
| **Anger** | Point | 63.33 | 16.67 | 0 | 0 | 16.67 | 3.33 |
| | Line | 63.33 | 13.33 | 3.33 | 0 | 16.67 | 3.33 |
| | Triangle | **70** | 6.67 | 10 | 0 | 10 | 3.33 |
| **Disgust** | Point | 23.33 | 56.67 | 3.33 | 13.33 | 0 | 3.33 |
| | Line | 6.67 | 66.67 | 10 | 16.67 | 0 | 0 |
| | Triangle | 10 | **80** | 3.33 | 6.67 | 0 | 0 |
| **Fear** | Point | 10 | 5 | 40 | 5 | 5 | 35 |
| | Line | 16 | 4 | 60 | 4 | 0 | 16 |
| | Triangle | 10 | 0 | **70** | 0 | 5 | 15 |
| **Happiness** | Point | 0 | 7.5 | 2.5 | 87.5 | 2.5 | 0 |
| | Line | 0 | 2.5 | 5 | **92.5** | 0 | 0 |
| | Triangle | 0 | 7.5 | 5 | 85 | 2.5 | 0 |
| **Sadness** | Point | 13.33 | 6.67 | 6.67 | 0 | **73.33** | 0 |
| | Line | 16.67 | 6.67 | 3.33 | 0 | **73.33** | 0 |
| | Triangle | 20 | 3.33 | 3.33 | 0 | **73.33** | 0 |
| **Surprise** | Point | 0 | 0 | 12.5 | 0 | 2.5 | 85 |
| | Line | 2.5 | 2.5 | 5 | 0 | 0 | **90** |
| | Triangle | 0 | 2.5 | 12.5 | 0 | 0 | 85 |

**Table 4.** Confusion matrix for FER in percentages using SVM classifier with different types of geometric features representation in MUG dataset.

| % | Feature Representation | Anger | Disgust | Fear | Happiness | Sadness | Surprise |
|---|---|---|---|---|---|---|---|
| **Anger** | Point | 96.36 | 0 | 1.82 | 0 | 1.82 | 0 |
| | Line | **100** | 0 | 0 | 0 | 0 | 0 |
| | Triangle | **100** | 0 | 0 | 0 | 0 | 0 |
| **Disgust** | Point | 0 | 98.18 | 0 | 0 | 1.82 | 0 |
| | Line | 1.82 | 96.36 | 0 | 1.82 | 0 | 0 |
| | Triangle | 0 | **100** | 0 | 0 | 0 | 0 |
| **Fear** | Point | 2.22 | 0 | 75.56 | 2.22 | 8.89 | 11.11 |
| | Line | 2.22 | 0 | 84.44 | 2.22 | 4.44 | 6.67 |
| | Triangle | 2.5 | 0 | **85** | 0 | 7.5 | 5 |
| **Happiness** | Point | 0 | 0 | 1.82 | 96.36 | 0 | 1.82 |
| | Line | 0 | 0 | 0 | **100** | 0 | 0 |
| | Triangle | 0 | 0 | 0 | **100** | 0 | 0 |
| **Sadness** | Point | 8 | 0 | 6 | 0 | 86 | 0 |
| | Line | 6 | 2 | 4 | 0 | 88 | 0 |
| | Triangle | 6 | 2 | 2 | 0 | **90** | 0 |
| **Surprise** | Point | 0 | 0 | 4 | 0 | 0 | 96 |
| | Line | 0 | 0 | 4 | 0 | 0 | 96 |
| | Triangle | 0 | 0 | 2 | 0 | 0 | **98** |

*6.2. Facial Expression Recognition using Boosted Line based Features*

As explained in section 3.3, the line is created by connecting two facial key points. With 52 facial key points 1326 unique lines are possible. But the features from only a subset of those lines are sufficient to learn the basic facial expressions. Therefore AdaBoost algorithm is used to select the discriminating lines from which features for SVM classification are extracted. The magnitude of difference in length and base angle w.r.t the neutral frame are extracted from line based features from facial expression sequence represented by Eq. (11) which is given as follows:

$$\Delta d^k_{(i,j),\max} = \max\left(abs(\Delta d^k_{1(i,j)}), abs(\Delta d^k_{2(i,j)}), ... abs(\Delta d^k_{N(i,j)})\right)$$
$$\Delta \theta^k_{(i,j),\max} = \max\left(abs(\Delta \theta^k_{1(i,j)}), abs(\Delta \theta^k_{2(i,j)}), ... abs(\Delta \theta^k_{N(i,j)})\right)$$

The average recognition accuracy using line based features with ten-fold cross validation is 96.58%, 74.31%, and 94.13% in CK+, MMI, and MUG dataset, respectively. There is improvement in recognition accuracy using line based features as compared to point based features. Table 2, 3 and 4 show the corresponding confusion matrices labeled with line based feature representation along with point and triangle based features.

*6.3. Facial Expression Recognition using Boosted Triangle based Features*

The overall procedure for extracting triangular features is explained in section 3.4. Multi-class AdaBoost with ELM is used to select the most discriminative features in the form of triangle composed of facial landmarks. A triangle in the proposed system, which is formed using facial key points, is represented by four components; two side lengths, an included angle, and a base angle of one of the triangle side with the x-axis (refer to fig. 4).

As shown in Fig. 4, features from triangle for SVM classification are extracted by subtracting triangle components of facial landmarks. The maximum changes in magnitude of the four components of the triangle in the sequence with respect to the triangle components in the first frame are extracted. Therefore each triangle is composed of four features, but some triangles in the AdaBoost selected triangle set may share the common edge, therefore the total feature dimension is always less or equal to the number of AdaBoost selected triangles multiplied by 4. As the number of triangle in the set increases the classification accuracy also increases. Fig. 8 shows the graph of the number of triangular features verses recognition accuracy in MUG facial expression dataset.

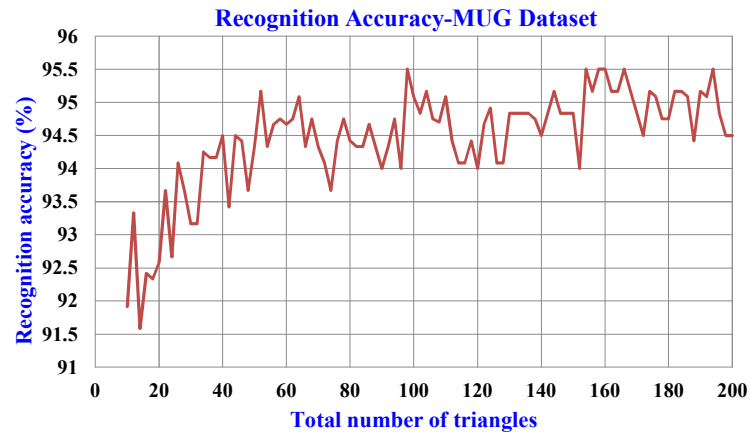

**Fig. 8.** Recognition accuracy under different number of boosted triangular features in MUG dataset

Table 2, 3 and 4 shows the confusion matrix for the FER using features extracted from 160, 84, and 98 AdaBoosted triangles in CK+, MMI, and MUG facial expression data sets respectively (labeled as triangle feature representation) along with point and line based features. The dimensionality of the feature vector using 160, 84 and 98 triangles in CK+, MMI and MUG dataset is 370, 317 and 330 respectively. The average recognition accuracies are 97.80%, 77.22% and 95.50% respectively.

We also performed the experiment by reducing the number of key points. As shows in Fig. 9, 25 and 34 key points tracking results are used to select the triangular features. The set, 25 key points, are the same set of key points used in [39] for the comparison of face recognition algorithms. Another set, 34 key points, are obtained by adding some more key points to the 25 key points set, especially in mouth and eyebrow regions. Finally, 52 key points are the set of key points used in [25], which is the extension of 34 key points set.

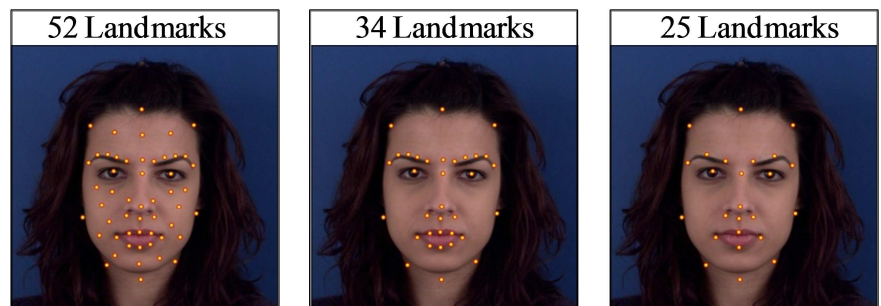

**Fig. 9.** Three different set of facial landmarks used for the evaluation of proposed triangular feature based FER system

Fig. 10 shows the comparison of FER accuracy in three different data sets using triangle based features with different number of facial key points tracking. In CK+ dataset 97.80%, 97.29%, and 93.37% of recognition accuracy, in MMI dataset 77.22%, 71.11%, and 68.61% of recognition accuracy and in MUG dataset 95.50%, 93.92%, and 93.00% of recognition accuracy is obtained using features from tracking result of 52, 34, and 25 facial landmarks respectively. Even if the small numbers of facial landmarks are used, good recognition accuracy can be obtained using the proposed triangle based geometric features.

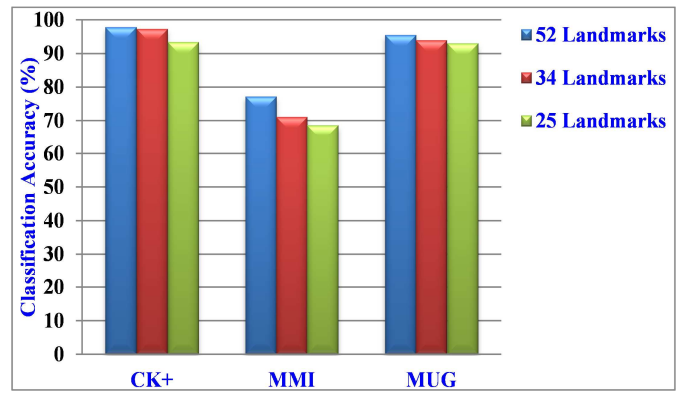

**Fig. 10.** Comparison of FER accuracy in three different data sets with triangular feature extracted from different set of landmarks tracking results

*6.4. Comparison of Point, Line and Triangle Feature based Facial Expression Recognition*

The features extracted based on the tracking result of single landmark is very simple. It gives the maximum displacement of feature point in four directions as some expression evolves over time. The second type of geometric feature is the features extracted based on line connecting two facial landmarks. Finally, third type of geometric feature is extracted in the form of components of triangles composed of facial key points.

Fig. 11 shows the comparison of FER performance using three different kinds of geometric features in three data sets. The average classification accuracy using point, line and triangle features in CK+ dataset is 96.37%, 96.58%, and 97.80%, in MMI dataset is 67.64%, 74.31%, and 77.22%, and in MUG dataset is 91.41%, 94.13%, and 95.50%, respectively. The features extracted in the form of line components give better result than point based features. On the other hand, feature extracted in the form of triangle components give better result than line based features. It proves that while some facial expression evolves, the movement of facial key points is not independent, i.e., there is some definite relationship between movement of facial key points.

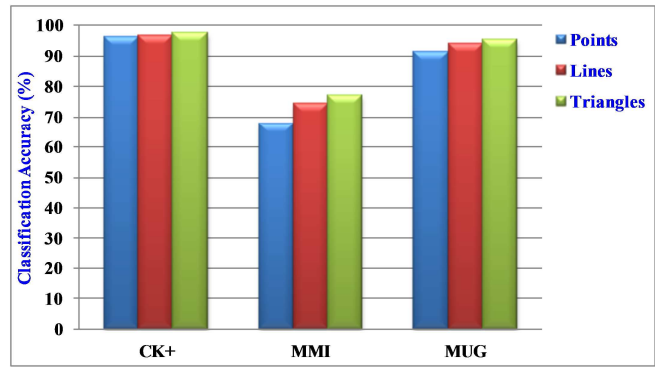

**Fig. 11.** Comparison of point, line and triangle feature based FER in three different data sets

The performance in the MMI dataset is low as compared to CK+ and MUG dataset. This is because MMI is difficult dataset among three data sets. Even though the line based feature give better result than point based feature, and triangle based feature is superior than line based feature, the performance in CK+ and MUG dataset using point, line, and triangle feature produce comparable results. But in conclusion, the best result in all three data sets is obtained using geometric features extracted based on the triangle composed of facial landmarks.

*6.5. Generalization Validation across Different Databases*

The generalization performance of the FER system can be best evaluated using cross-dataset evaluation. Most of the researchers use same dataset for both training and validation. It is obvious that higher recognition rate can be achieved when evaluated on a single dataset, because, while recording the facial expressions in lab environment, there is lot of similarity in all the recorded sequences. For example, lighting, background, way of expressing emotion, image quality, image resolutions etc. But if we take two different data sets, there will be much dissimilarity. The FER system will be a good one if it can produce better result when evaluated on a testing dataset different from the training dataset.

Table 5 shows the confusion matrix for FER using triangle based geometric feature while using CK+ dataset for training and MMI dataset for testing. The average recognition accuracy in this case is only 64.89%. Table 6 shows the confusion matrix for FER using MUG dataset for training and CK+ dataset for testing. In this case, the average recognition accuracy is 81.74%. Now, table 7 shows the cross-dataset evaluation result in three different dataset in which only average recognition result are presented. From these tables it can be seen that, while MMI dataset is used either for training or testing, the recognition accuracy is low, on the other hand if CK+ and MUG dataset are used for cross dataset evaluation the recognition performance is more than 80%. It shows that MMI dataset is the most difficult dataset among the three different dataset used in this paper. The cross-dataset evaluation is performed only for triangle based features because this feature set gives the best recognition accuracy as compared to point and line based features.

**Table 5.** Confusion matrixes for FER in percentages using SVM with boosted triangle based features while using CK+ dataset for training and MMI dataset for testing.

| %         | Anger | Disgust | Fear | Happiness | Sadness | Surprise |
|-----------|-------|---------|------|-----------|---------|----------|
| Anger     | **46.88** | 21.88 | 6.25 | 6.25 | 18.75 | 0 |
| Disgust   | 6.67 | **73.34** | 10 | 10 | 0 | 0 |
| Fear      | 7.14 | 10.71 | **50** | 3.57 | 21.43 | 7.14 |
| Happiness | 0 | 2.48 | 2.48 | **92.56** | 2.48 | 0 |
| Sadness   | 25.80 | 6.45 | 6.45 | 0 | **61.29** | 0 |
| Surprise  | 0 | 0 | 10 | 0 | 25 | **65** |

**Table 6.** Confusion matrixes for FER in percentages using SVM with boosted triangle based features while using MUG dataset for training and CK+ dataset for testing.

| %         | Anger | Disgust | Fear | Happiness | Sadness | Surprise |
|-----------|-------|---------|------|-----------|---------|----------|
| Anger     | **93.18** | 0 | 0 | 0 | 6.81 | 0 |
| Disgust   | 33.87 | **58.06** | 0 | 1.61 | 6.45 | 0 |
| Fear      | 0 | 11.11 | **55.56** | 14.81 | 18.52 | 0 |
| Happiness | 0 | 1.45 | 0 | **98.55** | 0 | 0 |
| Sadness   | 3.12 | 0 | 3.12 | 0 | **93.75** | 0 |
| Surprise  | 0 | 0 | 8.64 | 0 | 0 | **91.36** |

**Table 7.** Cross-dataset evaluation performance showing average recognition accuracies

| Training/Testing | MMI | CK+ | MUG |
|------------------|-----|-----|-----|
| MMI | X | 78.08 | 79.20 |
| CK+ | 64.89 | X | 83.09 |
| MUG | 64.56 | 81.74 | X |

*6.6. Comparison with state-of-the-art Methods*

Even though the experimental setup is not exactly same, the overall recognition accuracy of some recent methods of FER from the literature is compared with accuracy obtained using proposed systems. Geometric feature based FER system in which the features in the form of triangle are selected and the recognition is performed using SVM classification gives the best recognition accuracy of 97.8% in CK+ dataset, 77.22% in MMI dataset and 95.5% in MUG dataset.

In the literature so far, the system in [26] has shown superior performance, and has achieved 99.7% of recognition rate in CK+ dataset using key point displacement features. But in their method, the facial key point initialization is a manual process, and the number of key points is also larger than the number of key points used in the proposed method. On the other hand, the proposed system is fully automatic. Similarly, in [48], 97.16% recognition rate has been achieved by extracting the most discriminative facial key points for each facial expression. Recently in CK+ dataset, [31] have achieved 83.01% recognition accuracy in which geometric features are extracted using only 8 facial key points in the single highly expressed facial expression frame. In [14], using 96 image sequences from MMI dataset with LBP features; they achieved average recognition accuracy of 86.9%. Recently, Albert et al. [32] achieved 71.83% recognition accuracy in MMI dataset using attention theory based automatic sampling and optical flow as a temporal feature. In the proposed system 203 image sequences are used from MMI dataset, at which some of them are not acted facial expressions, i.e., they are naturally expressed facial expressions which adds difficulty in recognizing facial expressions with high accuracy. Rahulamathavan et al. [49] achieved 95.24% overall recognition accuracy in MUG facial expression dataset. They performed FER in encrypted domain using local fisher discriminant analysis. Recently, in [50], 92.76% and 94.31 % recognition accuracy is obtained in MUG and CK+ dataset using leave-one-subject-out validation strategy, respectively. The manifold structure is learned using coordinates of facial key points tracking result which can be decomposed to a small number of linear subspaces of very low dimension. Table 8 shows the summary of the comparison of FER performance with different methods in the literature. One of the advantages of the proposed geometric feature based FER systems is the relatively lower dimension of feature vector compared to the feature dimension in the state of the art methods. The recognition accuracy in CK+ and MUG dataset is comparable and even better than the best recognition accuracy in the literature. In the literature in MMI dataset the best accuracy is obtained using texture feature rather than geometric features.

**Table 8.** Comparison of FER performance with different methods in the literature.

| Reference | Method | Data sets | Class | Accuracy (%) |
|---|---|---|---|---|
| [26] | Semi-automatic, facial key point displacement features, SVM classifier | CK+ | 6 | 99.70 |
| [48] | Most discriminated facial key points for each facial expressions | CK+ | 6 | 97.16 |
| [31] | Geometric features from 8 facial key points, SVM classifier | CK+ | 7 | 83.01 |
| [14] | Boosted LBP features, SVM classifier | MMI | 7 | 86.90 |
| [32] | Attention theory based automatic sampling and optical flow as temporal features | MMI | 6 | 71.83 |
| [49] | Local fisher discriminant analysis in encrypted domain | MUG | 7 | 95.24 |
| [50] | Manifold structure learning using coordinates of facial key point tracking results | CK+ | 6 | 94.31 |
| [50] | | MUG | 6 | 92.76 |
| [21] | Graph-preserving sparse NMF | CK+ | 6 | 94.30 |
| [38] | Enhanced independent component, FLDA | CK+ | 6 | 93.23 |
| [30] | Geometric features, dynamic Bayesian network | CK+ | 6 | 94.04 |
| **Ours** | **Fully-automatic, triangle based geometric feature representation, salient feature selection, SVM classifier** | **CK+** | **6** | **97.80** |
| | | **MMI** | **6** | **77.22** |
| | | **MUG** | **6** | **95.50** |

## 7. Conclusion

The aim of this paper is to present a new framework for FER in frontal image sequence based on geometric features extracted from the tracking result of facial key points. Different types of geometric feature extraction techniques from facial expression image sequence are presented. The facial expressions are recognized using most discriminative geometric feature selected using feature selective AdaBoost algorithm.

The point, line and triangle based features are presented. The point based feature can be used directly, whereas line and triangle based feature are used only after feature selection process. The performance of the proposed geometric-feature based FER system is evaluated in three different data sets: namely CK+, MMI and MUG. The line based feature gives better result than point based feature, whereas triangle based feature gives superior result than line based features. Therefore recognition accuracy using the features extracted considering more key points at a time is better than using the features extracted by considering single key point at a time. Therefore the most desirable feature is the time-varying graph itself. But we cannot use graph directly, so we need to find out efficient feature from it, which do not reduce the information in the graph.

The recognition accuracy in CK+ and MUG dataset is more than 95%, whereas in MMI dataset the recognition accuracy is only 77.22%. The MMI dataset is relatively difficult then CK+ and MUG dataset because it includes some spontaneous facial expressions. The generalization capability of the proposed FER system is proved using cross-dataset evaluation. More than 80% recognition accuracy is obtained using proposed FER system while using different data sets for training and testing. While comparing the results with the state of the art methods, the performances of the proposed system is comparable and at times even better than the results reported in the literature for most cases.